# A Revision-Based Approach to Resolving Conflicting Information

**Guilin Qi**
School of Computer Science
Queen's University Belfast
Belfast, BT7 1NN, UK
G.Qi@qub.ac.uk

**Weiru Liu**
School of Computer Science
Queen's University Belfast
Belfast, BT7 1NN, UK
W.Liu@qub.ac.uk

**David A. Bell**
School of Computer Science
Queen's University Belfast
Belfast, BT7 1NN, UK
DA.Bell@qub.ac.uk

## Abstract

In this paper, we propose a revision-based approach for conflict resolution by generalizing the *Disjunctive Maxi-Adjustment* (DMA) approach (Benferhat et al. 2004). Revision operators can be classified into two different families: the model-based ones and the formula-based ones. So the revision-based approach has two different versions according to which family of revision operators is chosen. Two particular revision operators are considered, one is the Dalal's revision operator, which is a model-based revision operator, and the other is the cardinality-maximal based revision operator, which is a formula-based revision operator. When the Dalal's revision operator is chosen, the revision-based approach is independent of the syntactic form in each stratum and it captures some notion of minimal change. When the cardinality-maximal based revision operator is chosen, the revision-based approach is equivalent to the DMA approach. We also show that both approaches are computationally easier than the DMA approach.

## 1 Introduction

Inconsistency handling is a fundamental problem in artificial intelligence. This problem has been addressed in several important areas, such as knowledge integration, iterated belief revision and exception handling. Many approaches to handling inconsistency have been proposed (Amgoud and Cayrol 2002, Benferhat et al. 1993, Benferhat et al. 2004, Coste-Marquis and Marquis 2000, Nitta 04, Williams 94, Williams 96). Among them, an important class of approaches is *coherence-based*, which selects one (several) consistent subbase(s) and then applies the classical inference on this (these) subbase(s). In most of the *coherence-based approaches*, pieces of information are represented by propositional stratified knowledge[1] bases, i.e finite sets of propositional formulae equipped with a total pre-order which represents the available preferences over the given beliefs.

Recently, several approaches based on some kinds of *Adjustment* procedures are discussed, which compute a consistent knowledge base by removing part of conflicting information or weakening conflicting information. Williams firstly proposed to deal with conflicts in knowledge bases by so-called *Adjustment* system (Williams 94). Later, she introduced another strategy called *Maxi-Adjustment* to improve *Adjustment*. In (Benferhat et al. 2004), an approach, called *Disjunctive Maxi-Adjustment* (DMA for short), was proposed. Both *Maxi-Adjustment* and DMA solve conflicts at each level of priority in the knowledge base. Starting from the information with the lowest stratum where formulae have highest level of priority, when inconsistency is encountered in the knowledge base, *Maxi-Adjustment* removes all formulae in the higher strata responsible for the conflicts, whilst DMA weakens the conflicting information in those strata. A very important property of DMA is that it is a compilation of the lexicographical system (Benferhat et al. 1993). DMA needs to compute all the conflicts in a stratum, which is a hard task. Therefore two alternative implementations of DMA are given to reduce the computational complexity. One is called whole-DMA and the other is called iterative-DMA. Whole-DMA is computationally much more tractable than both DMA and iterative-DMA because it does not need to compute *conflicts*. However, the size of the knowledge base obtained by whole-DMA is larger than the original one exponentially in the worst case.

In this paper, we propose a revision-based approach for conflict resolution by generalizing the DMA approach.

---
[1] We use the terms *belief* and *knowledge* interchangeably in this paper.

Similar to the DMA approach, we assume that a new sure formula $\phi$ is added to some stratified knowledge base (this knowledge base may be inconsistent). We start by revising the set of formulae in the second stratum using $\phi$, then we revise the set of formulae in the third stratum using $\phi$ and the formulae kept in the second stratum after revision, and so on.

Revision operators can be classified into two different families: the model-based ones and the formula-based ones. So the revision-based approach has two different versions according to which family of revision operators is chosen. Two particular revision operators are considered, one is the Dalal's revision operator, which is a model-based revision operator, and the other is the cardinality-maximal based revision operator, which is a formula-based revision operator. When the Dalal's revision operator is chosen, the revision-based approach is independent of the syntactic form in each stratum and it captures some notion of minimal change. When the cardinality-maximal based revision operator is chosen, the revision-based approach is equivalent to the DMA approach. We also show that both approaches are computationally easier than the DMA approach.

This paper is organized as follows. Section 2 gives a brief review of stratified knowledge bases. We then give a glimpse at two important propositional knowledge base revision operators in Section 3. Section 4 reviews the DMA approach and two of its alternative implementations. In Section 5, we present our revision-based approach to handling inconsistency. In Section 6, two particular revision-based approaches are given. Finally, we conclude the paper in Section 7.

## 2 Stratified Knowledge Bases

In this paper, we consider a propositional language $\mathcal{L}$ over a finite alphabet $\mathcal{P}$. An interpretation is a truth assignment to the atoms in $\mathcal{P}$, i.e. a mapping from $\mathcal{P}$ to $\{true, false\}$. We denote the set of classical interpretations by $\Omega$, propositional variables by $a, b, ...$, and classical formulae by $\phi, \psi, \gamma, ...$. Capital letters $A, B, C, ...$ represent sets of classical formulae. Deduction in classical propositional logic is denoted by the symbol $\vdash$ and a deductive closure by $Cn$ such that $Cn(A) = \{\phi | A \vdash \phi\}$. An interpretation $\omega$ is a model of a formula $\phi$ if $\phi$ evaluates to $true$ in $\omega$. We use $Mod(\phi)$ to denote the set of models of $\phi$. Conversely, let $M$ be a set of interpretations, $form(M)$ denotes the logical formula (unique up to logical equivalence) whose models are $M$. A *(flat) knowledge base (KB for short)* $K$ is a finite set of propositional formulae which can be seen as a formula $\phi$ which is the conjunction of the formulae of $K$. If $\mathcal{S} = \{K_1, ..., K_m\}$ is a finite family of finite knowledge bases, then $\bigvee \mathcal{S} = \{\phi_1 \vee ... \vee \phi_m : \phi_i \in K_i\}$ and $\bigcap \mathcal{S} = \cap \{A : A \in \mathcal{S}\}$. As usual, we set $\bigvee \emptyset = \bot$.

A *stratified* knowledge base is a set of formulae with the form $K = (S_1, ..., S_n)$, where $S_i$ ($i = 1, ..., n$) is a stratum containing propositional formulae of $K$ having the same rank or level of priority such that each formula in $S_i$ is more reliable than formulae of the stratum $S_j$ for $j > i$. Namely, the lower the stratum, the higher the rank. The importance of the stratified knowledge bases has been addressed in many AI areas (Amgoud and Cayrol 2002, Benferhat et al. 1993, Benferhat et al. 2004, Williams 1994, Williams 1996). The priority relation of a stratified knowledge base makes it easier to deal with inconsistency. Given a stratified knowledge base $K = (S_1, ..., S_n)$, a conflict in $K$, denoted by $C$, is a subbase of $K$ such that $C \vdash \bot$ and $\forall C' \subset C, C' \nvdash \bot$. The *kernel* of $K$ is the union of all its conflicts, that is, it is the set of formulae of $K$ which are involved in at least one conflict.

## 3 Propositional Knowledge Base Revision

In their pioneer work, Gädenfors and his colleagues proposed a set of rational postulates, known as AGM postulates, to characterize a belief revision operator (Gärdenfors 1988). Instead of a finite KB, they consider a *knowledge set*, which is a set of formulae closed under deduction. However, it is representationally infeasible to model belief states by a knowledge set in a computer-based application because logically closed sets are always infinite. Many propositional knowledge base revision operators have been proposed (see (Eiter and Gottlob 1992, Katsuna and Mendelzon 1991, Nebel 1998) for a full list of them). There are two different families of revision operators: the model-based and the formula-based. In this section, we first review two important revision operators. The first one is Dalal's revision operator $\circ_D$ (Dalal 1988), which is a model-based revision operator, and the other is the cardinality-maximizing base (CMB) revision operator $\circ_C$, which is a formula-based revision operator (Nebel 1998). We then propose a revision operator by revising the CMB revision operator.

Dalal (Dalal 1988) first defines a measure of "distance" $dist(\omega_1, \omega_2)$ between two interpretations $\omega_1$ and $\omega_2$ as the number of propositional letters on which they differ. The distance between a knowledge base $K$ and an interpretation $\omega$ is defined as

$$dist(K, \omega) = min_{\omega_i \in Mod(K)} dist(\omega, \omega_i).$$

He then defines a total pre-order $\leq_K$ as $\omega_1 \leq_K \omega_2$ if and only if

$$dist(Mod(K), \omega_1) \leq dist(Mod(K), \omega_2).$$

As usual, we define $\omega_1 < \omega_2$ as $\omega_1 \leq \omega_2$ but $\omega_2 \not\leq \omega_1$.

Finally, Dalal's revision operator $\circ_D$ can be defined as: given a KB $K$ and a formula $\mu$, the revision of $K$ by $\mu$ is
$$Mod(K \circ_D \mu) = min(Mod(\mu), \leq_K).$$

So the result of revision of $K$ by $\mu$ by Dalal's revision operator consists of the "minimal" models of $\mu$ with regard to the total pre-order $\leq_K$.

Now let us introduce CMB-revision operator.

Let $(K \perp \phi)$ be the cardinality-maximal subsets of $K$ that are consistent with $\neg \phi$, i.e.,

$K \perp \phi = \{A \subseteq K | A \not\models \phi, \forall B \subseteq K, \text{ if } |A| < |B|, \text{ then } B \models \phi\}$,

where $|A|$ denotes the cardinality of the set $A$.

The cardinality-maximizing base revision is defined as follows (Nebel 1991):
$$K \circ_{CMB} \phi = Cn(\bigvee ((K \perp \neg \phi)) \cup \{\phi\}).$$

The result of CMB-revision is a knowledge set, so it has the same problem as before, i.e it is representationally infeasible in computer-based application. So we propose another revision operator, called cardinality-maximizing based revision operator $\circ_{CM}$, which is defined as follows:
$$K \circ_{CM} \phi = \bigvee ((K \perp \neg \phi)) \cup \{\phi\}.$$

The operator $\circ_{CM}$ will be used to define a particular revision-based approach in Section 6.

Let us look at the computational complexity of CM-revision.

**Proposition 1** *Generating a revised base under CM-revision is in $F\Delta_2^p$, where $\Delta_2^p$ denotes the set of decision problems decidable by a polynomial-time Turing machine with an NP oracle, and "F" in $F\Delta_2^p$ stands for* function *and is intended to turn a complexity class for decision problem into one for* search *problem.*

Proposition 1 can be proved by considering the proof of Theorem 5.14 in (Nebel 1998).

$K \circ_{CM} \phi$ defined above may contain some redundant information.

**Example 1** *Let $K = \{a, b, c, d\}$ and $\phi = \neg a \vee \neg b \vee \neg c$. Then $K \perp \neg \phi = \{\{a, b, d\}, \{a, c, d\}, \{b, c, d\}\}$. So $K \circ_{CM} \phi = \bigvee (K \perp \neg \phi) \cup \{\phi\} = \{a \vee b, a \vee c, b \vee c, a \vee b \vee c, d, a \vee d, b \vee d, c \vee d, a \vee b \vee d, a \vee c \vee d, b \vee c \vee d, \phi\}$, which is equivalent to $\{a \vee b, a \vee c, b \vee c, d, \phi\}$.*

**Definition 1** *A disjunction $\phi = \phi_1 \vee ... \vee \phi_n$ is subsumed by disjunction $\psi = \psi_1 \vee ... \vee \psi_m$, denoted as $\psi \sqsubseteq \phi$ iff $\{\psi_1, ..., \psi_m\} \subseteq \{\phi_1, ..., \phi_n\}$*

We compute $K \circ_{CM} \phi$ in the following way.

First, suppose $C = \bigcap (K \perp \neg \phi)$, then we can define
$$K' = \bigvee \{A \backslash C : A \in K \perp \neg \phi\} \cup C \cup \{\phi\}.$$

Let $D = \bigvee \{A \backslash C : A \in K \perp \neg \phi\}$. Some disjunctions in $D$ can be subsumed by other elements in it. Let us define
$$K_{CM} = \{\phi_i \in D : \nexists \psi \in D, \psi \sqsubseteq \phi_i\} \cup C \cup \{\phi\}.$$

It is easy to check that $K_{CM} \equiv K \circ_{CM} \phi$. Moreover, since we can decide whether a disjunction is subsumed by another one in polynomial time, the complexity of computing $K_{CM}$ is in the same level of the polynomial hierarchy as that of computing $K \circ_{CM} \phi$.

## 4 Disjunctive Maxi-Adjustment (DMA)

In this section, we describe the Disjunctive Maxi-Adjustment (DMA) and its two alternative implementations whole DMA and iterative DMA from (Benferhat et al. 2004). In the following, we use $K_\phi$ to denote a stratified knowledge base $\{\phi\} \cup K$, that is, we add to $K$ ($K$ is assumed to be consistent) a stratum with the highest rank which consists of a single formula $\phi$. Let $d_k(C)$ be the set of all possible disjunctions of size $k$ between formulae of $C$. If $k > |C|$ then $d_k(C) = \emptyset$.

**Algorithm 1:** DMA $(K, \phi)$

Data: a stratified knowledge base $K = \{S_1, ..., S_n\}$; a new sure formula $\phi$;

Result: a consistent subbase $\delta_{DMA}(K_\phi)$

**begin**

$KB \leftarrow \{\phi\}$;

**for** $i = 1$ to $n$ **do**

    **if** $KB \cup S_i$ is consistent **then** $KB \leftarrow KB \cup S_i$

    **else**

        Let $C$ be the subset of $S_i$ in kernel of $KB \cup S_i$;

        $KB \leftarrow KB \cup \{\phi : \phi \in S_i \text{ and } \phi \notin C\}$;

        $k \leftarrow 2$;

        **while** $k \leq |C|$ and $KB \cup d_k(C)$ is inconsistent

        **do**

            $k \leftarrow k + 1$;

        **if** $k \leq |C|$ **then** $KB \leftarrow KB \cup d_k(C)$;

**return** $KB$

**end**

The idea of DMA is that we start from the first stratum and take the formulae of $S_1$ which do not belong to any conflict in $\{\phi\}\cup S_1$. For those formulae involved in the conflicts at this stratum we weaken them by replacing them with their pairwise disjunctions. If the result is consistent at this stratum we move to the next stratum, else we replace these formulae by their possible disjunctions of size 3, and so on.

**Example 2** *Let $\phi = c$ and $K = \{S_1, S_2, S_3\}$ be such that $S_1 = \{a\vee b\}$, $S_2 = \{\neg a, \neg b, \neg c\vee b, d, e\}$ and $S_3 = \{\neg c \vee \neg d\}$. First we have $KB = \{c\}$. There is no conflict in $KB\cup S_1$ then $KB\leftarrow\{a\vee b, c\}$. Now, $S_2$ contradicts $KB$ due to the conflicts $\{a\vee b, \neg a, \neg b\}$ and $\{c, \neg b, \neg c\vee b\}$. Since $d, e$ are not involved in any conflict, so $KB\leftarrow KB\cup\{d, e\}$. Now we create all the possible disjunctions of size 2 with $C = \{\neg a, \neg b, \neg c\vee b\} : d_2(C) = \{\neg a \vee \neg b, \neg a \vee \neg c\vee b\}$. Since $KB\cup d_2(C)$ is consistent, we add $d_2(C)$ to $KB$, i.e. $KB\leftarrow KB\cup d_2(C)$. Finally, since $KB\cup S_3$ is inconsistent, and we cannot create larger disjunctions because $S_3$ contains only a single formula, we do not add anything from $S_3$ to $KB$ and the algorithm stops. Then $\delta_{DMA}(K_\phi) = \{a\vee b, c, \neg a \vee \neg b, \neg a \vee \neg c\vee b, d, e\}$, which is equivalent to*

$$\delta_{DMA}(K_\phi) = \{\neg a, b, c, d, e\}.$$

A disadvantage of DMA is that it needs to compute the kernel, which is in general a hard problem (Bessant et al. 2001). So two modified versions of the DMA algorithm are proposed in (Benferhat et al. 2004) to reduce its computational complexity. One is called whole-DMA, which does not compute the kernel when $KB\cup S_i$ is inconsistent, instead, all possible disjunctions of size $j$ of $S_i$ are considered. Therefore, it is computationally much more easier than DMA. However, it produces a large number of disjunctions, that is, the size of the resulting knowledge base may be exponentially larger than the original one. The other one is called iterative-DMA, which still needs to compute the kernel. Suppose $KB\cup S_i$ is inconsistent, we then compute $d_2(C)$, where $C$ is the kernel of $S_i$. When $KB\cup(S_i \setminus C) \cup d_2(C)$ is still inconsistent, then rather than weakening $C$ again by considering disjunctions of size 3, we only weaken those formulae in $d_2(C)$ which are still responsible for conflict. It has been shown that the knowledge base obtained by DMA is equivalent to those of both whole-DMA and iterative-DMA.

## 5 Revision-based Approach to Handling Inconsistency

In this section, we propose an approach to handling inconsistency which is based on a revision operator. The idea is that we start by revising the set of formulae in the second stratum using the set of formulae in the first stratum, then we revise the third stratum using the set of formulae in the first stratum and formulae kept in the second stratum after revision, and so on. since there are two families of revision operators, our revision-based approach has two different versions according to which family of revision operators is chosen. In this section and the following sections, the knowledge base $K = \{S_1, ..., S_n\}$ may be inconsistent but each $S_i$ must be consistent.

**Revision-based Algorithm I: model-based revision operator**

Input: a stratified knowledge base $K = \{S_1, ..., S_n\}$; a new sure formula $\phi$; a model-based revision operator $\circ$;

Result: a set of models of $\delta_{Rev}(K_\phi)$

**begin**

$KB\leftarrow\{\phi\}$;

**for** $i = 1$ to $n$ **do**

    **if** $Mod(KB)\cap Mod(S_i)\neq\emptyset$ **then** $Mod(KB)\leftarrow Mod(KB)\cap Mod(S_i)$

    **else**

        Let $\psi = Form(Mod(KB))$

        $Mod(KB)\leftarrow Mod(S_i\circ\psi)$;

**return** $Mod(KB)$

**end**

Since the revision-based Algorithm I uses a model-based revision operator to resolve the conflicting information in each stratum, it is irrelevant of the syntactical form of the set of formulae in each stratum. That is, given two stratified knowledge bases $K = \{S_1, ..., S_n\}$ and $K' = \{S'_1, ..., S'_n\}$, and two formulae $\phi$ and $\phi'$, if $S_i\equiv S'_i$ and $\phi\equiv\phi'$, then $Mod(\delta_{Rev}(K_\phi)) = Mod(\delta_{Rev}(K'_{\phi'}))$.

Next we define another revision based algorithm where a formula-based revision operator is applied.

**Revision-based Algorithm II: formula-based revision operator**

Input: a stratified knowledge base $K = \{S_1, ..., S_n\}$; a new sure formula $\phi$; a formula-based revision operator $\circ$;

Result: a consistent subbase $\delta_{Rev}(K_\phi)$

**begin**

$KB\leftarrow\{\phi\}$;

```
for i = 1 to n do
    if KB∪S_i is consistent then KB←KB∪S_i
    else
        Let ψ = ∧{φ_i : φ_i∈KB}
        KB←S_i∘ψ;
return KB
end
```

The revision-based Algorithm II returns a knowledge base. Since it resolves inconsistency in each stratum using a formula-based revision operator, the syntactical form of each stratum will influence the output of the algorithm.

**Definition 2** *Let $K$ be a classical knowledge base and $\phi$ be a new formula. Let $C$ be the subset of $K$ in kernel of $K\cup\{\phi\}$. Then the DMA revision operator, denoted as $\circ_{DMA}$, is defined as:*

$$K\circ_{DMA}\phi = d_k(C)\cup F \cup \{\phi\},$$

*where $k$ is the minimal natural number $i$ such that $d_i(C)\cup\{\phi\}$ is consistent and $F = K \setminus C$.*

Clearly, when the DMA revision operator is chosen, the revision-based Algorithm II is reduced to Algorithm 1. So our revision-based Algorithm II generalizes Algorithm 1.

Our revision-based algorithms can be used to deal with belief revision and knowledge integration where the original knowledge base(s) is (are) stratified and inconsistent. In belief revision, the original knowledge base is usually assumed to be consistent. This assumption is criticized by some researchers (Priest 2001, Tanaka 2005). Until now, little work has been done in belief revision where the original knowledge base is both stratified and inconsistent. The revision methods derived by the revision-based algorithm can partially fill this gap.

## 6 Two Particular Cases

### 6.1 Case 1: Dalal's revision operator

Dalal's revision operator is one of the most important model-based revision operators in the literature. It satisfies the so-called AGM postulates which constrain the revision process so that minimal changes occur in the knowledge base. In this section, we will consider the revision-based algorithm I where the Dalal's revision operator is chosen, we call it Dalal's revision-based algorithm (DR). We use $\delta_{DR}$ to denote the resulting knowledge base of the DR approach.

Let us look at an example to illustrate DR.

**Example 3** *(Continue Example 2)* First we have $KB = \{c\}$. There is no conflict in $KB\cup S_1$, so $Mod(KB)\leftarrow\{\{a,c\},\{b,c\},\{a,b,c\}\}$. $KB\cup S_2$ is inconsistent. Since $Form(Mod(KB)) = (a\vee b) \wedge c$, we set $\psi = (a\vee b) \wedge c$. The revision of $S_2$ by $\psi$ is $Mod(S_2\circ_D\psi) = \{\{a,c,d,e\},\{b,c,d,e\}\}$. So $KB = \{(a\wedge\neg b\wedge c\wedge d\wedge e) \vee (\neg a\wedge b\wedge c\wedge d\wedge e)\}$. $KB\cup S_3$ is inconsistent. Since $Form(KB) = (a\vee b)\wedge(\neg a \vee \neg b)\wedge c\wedge d\wedge e$, we set $\psi = (a\vee b)\wedge(\neg a \vee \neg b)\wedge c\wedge d\wedge e$. The revision of $S_3$ by $\psi$ is $Mod(S_3\circ_D\psi) = \{\{a,c,d,e\},\{b,c,d,e\}\}$. So $Mod(\delta_{DR}(K_\phi)) = \{\{a,c,d,e\},\{b,c,d,e\}\}$.

In Example 1, the result of DMA is $\delta_{DMA}(K_\phi) = \{\neg a,b,c,d,e\}$, so $\delta_{DMA}(K_\phi) \vdash \delta_{DR}(K_\phi)$. However, the result of DMA does not always infer more information than that of DR, which can be seen from the following example.

**Example 4** *Let $K = \{S_1,S_2\}$ be such that $S_1 = \{a,c,d\}$ and $S_2 = \{\neg a \vee b, \neg b, d\rightarrow r, r\rightarrow\neg a\}$ and $\phi = c$. Let us first apply DR to handle inconsistency. First we have $KB = \{c\}$. There is no conflict in $KB\cup S_1$, then $Mod(KB)\leftarrow\{\{a,b,c,d\},\{a,\neg b,c,d\}\}$. $KB\cup S_2$ is inconsistent. Since $Form(Mod(KB)) = a\wedge c\wedge d$, we set $\psi = a\wedge c\wedge d$. The revision of $S_2$ by $\psi$ is $Mod(S_2\circ_D\psi) = \{\{a,c,d,r\}\}$. So $\delta_{DR}(K_\phi) = \{\{a,c,d,r\}\}$. Now let us look at DMA. First we have $KB = \{c\}$. There is no conflict in $KB\cup S_1$, then $KB\leftarrow\{a,c,d\}$. $KB\cup S_2$ is inconsistent. We now create all the possible disjunctions of size 2 with $C = \{\neg a \vee b, \neg b, d\rightarrow r, r\rightarrow\neg a\}$: $d_2(C) = \{\neg a\vee b\vee\neg d\vee r, \neg a\vee b\vee\neg r, \neg b\vee\neg d \vee r, \neg a \vee \neg b \vee \neg r\}$. Since $KB\cup d_2(C)$ is inconsistent, we create $d_3(C) = \{\top\}$. It is clear $KB\cup d_3(C)$ is consistent, so the algorithm stops and $\delta_{DMA}(K_\phi) = \{a,c,d\}$. Clearly we have $\delta_{DR}(K_\phi) \vdash \delta_{DMA}(K_\phi)$ but $\delta_{DMA}(K_\phi) \not\vdash \delta_{DR}(K_\phi)$.*

Let us define a revision operator $\circ_{DR}$ by $K\circ_{DR}\phi = \delta_{DR}(K_\phi)$, then we have the following properties for $\circ_{DR}$.

**Proposition 2** *The revision operator $\circ_{DR}$ satisfies the following properties.*

(D1) $K\circ_{DR}\phi$ *is satisfiable.*

(D2) $K\circ_{DR}\phi \vdash \phi.$

(D3) *If $K\cup\{\{\phi\}\}$ is consistent, then $K\circ_{DR}\phi\equiv K\cup\{\{\phi\}\}$.*

(D4) *Given two stratified knowledge bases $K = \{S_1,...,S_n\}$ and $K' = \{S'_1,...,S'_n\}$, and two formulae $\phi$ and $\phi'$, if $S_i\equiv S'_i$ and $\phi\equiv\phi'$, then $Mod(K\circ_{DR}\phi) = Mod(K'\circ_{DR}\phi')$.*

(D5) $(K \circ_{PIR} \phi) \wedge \psi$ implies $K \circ_{PIR}(\phi \wedge \psi)$.

(D6) If $(K \circ_{PIR} \phi) \wedge \psi$ is satisfiable, then $K \circ_{PIR}(\phi \wedge \psi)$ implies $(K \circ_{PIR} \phi) \wedge \psi$.

(D1)-(D3) correspond to Conditions (R1)-(R3) for a revision operator proposed in (Katsuno and Mendelzon 1991). (D4) is a generalization of Dalal's Principle of Irrelevance of Syntax (it is the condition (R4) in (Katsuno and Mendelzon 1991) to the stratified knowledge base. (D5) and (D6) are generalization of (R5) and (R6) in (Katsuna and Mendelzon 1991) which are important to ensure minimal change.

Another important property of $\circ_{DR}$ is that it captures some notion of minimal change.

Let $K = \{S_1, ..., S_n\}$ be a stratified knowledge base. Let us define a pre-order $\leq_K$ as $\omega_1 \leq_{K,Lex} \omega_2$ iff $\exists i$ $\omega_1 <_{S_i} \omega_2$ and $\forall j < i$: $\omega_1 =_{S_j} \omega_2$, where $\omega_1 =_{S_j} \omega_2$ iff $\omega_1 \leq_{S_j} \omega_2$ and $\omega_2 \leq_{S_j} \omega_1$.

**Proposition 3** *Let $K = \{S_1, ..., S_n\}$ be a stratified knowledge base and $\phi$ be a new sure formula. Suppose $K \circ_{DR} \phi$ is the result of revising $K$ by $\phi$ using $\circ_{DR}$, then*

$$Mod(K \circ_{DR} \phi) = min(Mod(\phi), \leq_{K,Lex}).$$

Proposition 3 shows that the revision operator $\circ_{DR}$ generalizes the Dalal's revision operator by generalizing the total pre-order $\leq_K$ to $\leq_{K,Lex}$.

By the proof of Theorem 6.9 in (Eiter and Gottlob 1992), generating a revised base under Dalal's revision operator is $FP^{NP[\mathcal{O}(logn)]}$, i.e it needs at most $[\mathcal{O}(logn)]$ calls to a $NP$ oracle to generate a revised base. In contrast, in each weakening step of DMA, it needs to compute kernel, which is a very hard problem because determining if a given formula $\phi$ is in the kernel of a knowledge base $K$ is $\Sigma_2^p$-compelete (Bessant et al. 2001). Therefore, the DMA approach is computationally harder than the DR approach (under the usual assumptions of complexity theory (Johnson 1990)).

### 6.2 Case 2: Cardinality-maximizing based revision (CM-revision) operator

In this section, we consider the cardinality-maximizing based revision (CM-revision for short) operator in revision-based algorithm II. We will prove that CM-revision-based algorithm (**CMR**) is equivalent to DMA and it is computationally easier than DMA.

**Definition 3** *A formula $\psi$ is said to be a CMR consequence of $K$ and $\phi$, denoted by $K_\phi \vdash_{CMR} \psi$ iff $\delta_{CMR}(K_\phi) \vdash \psi$, where $\delta_{CMR}(K_\phi)$ is the resulting knowledge base of the CMR approach.*

To prove that CM-revision-based algorithm is equivalent to DMA, we first prove that it is equivalent to the lexicographical system, which has been shown to be equivalent to DMA in (Benferhat et al. 2004).

**Definition 4** *(Benferhat et al. 2004) Let $K = \{S_1, ..., S_n\}$ be a stratified knowledge base. Let $A = \{A_1, ..., A_n\}$ and $B = \{B_1, ..., B_n\}$ be two consistent subbases of $K$. $A$ is said to be lexicographically preferred to $B$, denoted to $A >_{Lex} B$, iff*

$$\exists k \text{ s.t. } |A_k| > |B_k| \text{ and } \forall 1 \leq j < k, |A_j| = |B_j|.$$

Let $\delta_{Lex}(K)$ denote the set of all subbases of $K$ which are maximal with regard to $>_{Lex}$. Then, the lexicographical consequence is defined as follows.

**Definition 5** *(Benferhat et al. 2004) A formula $\psi$ is said to be a lexicographical consequence of $K_\phi$, denoted by $K_\phi \vdash_{Lex} \psi$, if it is a classical consequence of all the elements of $\delta_{Lex}(K_\phi)$, namely*

$$\forall A \in \delta_{Lex}(K_\phi), \quad A \vdash \psi.$$

We now give a theorem to show that CMR is equivalent to the lexicographical system.

**Lemma 1** *Let $K = \{S_1, ..., S_n\}$ be a stratified knowledge base, $\phi$ be a new formula, and $\delta_{Lex}(K_\phi) = \{A_1, ..., A_m\}$, where $A_i = \{\{\phi\}, A_{i1}, ..., A_{in}\}$. Then $\{\phi\} \cup \bigvee(\{B_i\}_{i=1,...,m}) \equiv \{\phi\} \cup \bigvee(\{A_{i1}\}_{i=1,...,m}) \cup .. \cup \bigvee(\{A_{in}\}_{i=1,...,m})$, where $B_i = A_{i1} \cup ... \cup A_{in}$.*

**Lemma 2** *Let $K = \{S_1, ..., S_n\}$ be a stratified knowledge base, $\phi$ be a new formula, and $\delta_{Lex}(K) = \{A_1, ..., A_m\}$, where $A_i = \{\{\phi\}, A_{i1}, ..., A_{in}\}$. Let $\delta_{CMR}(K_\phi)$ be the knowledge base obtained by CMR. Then $\delta_{CMR}(K_\phi) = \{\phi\} \cup \bigvee(\{A_{i1}\}_{i=1,...,m}) \cup .. \cup \bigvee(\{A_{in}\}_{i=1,...,m})$.*

**Proof:** *We prove Lemma 2 by induction over rank $i$ of $K$.*

*When $i = 1$, we know that $A_{11}, ..., A_{m1}$ are the set of cardinality-maximal subset of $\{\phi\} \cup S_1$. Suppose $KB$ is the knowledge base obtained by applying CMR algorithm in the first rank of $K$, we have $\delta_{CMR}(K_\phi) = \{\phi\} \cup \bigvee(\{A_{i1}\}_{i=1,...,m})$.*

*Now suppose Lemma 2 holds when $i < k$, we prove that it holds when $i = k$. Let $KB$ be the knowledge base obtained by CMR algorithm until stratum $i$-1. By assumption, we have $KB = \{\phi\} \cup \bigvee(\{A_{i1}\}_{i=1,...,m}) \cup .. \cup \bigvee(\{A_{i,k-1}\}_{i=1,...,m})$. We now need to prove that the set of cardinality-maximal consistent subsets of $KB \cup S_k$ is $\{KB \cup A_{ik}\}_{i=1,...,m}$. First, suppose $A \subseteq S_k$ such that $KB \cup A$ is a cardinality-maximal subset of $KB \cup S_k$, then there must exist $A_{i_1,1}, ..., A_{i_{k-1},k-1}$*

such that $\{\phi\}\cup A_{i_1,1} \cup ... \cup A_{i_{k-1},k-1}\cup A$ is a consistent subset of $\{\phi\}\cup S_1...\cup S_k$, where $A_{i_j,j}\in\{A_{ij}\}_{i=1,...,m}$. Moreover, $\{\phi\}\cup A_{i_1,1} \cup ...\cup A_{i_{k-1},k-1}\cup A$ must be a cardinality-maximal subset of $\{\phi\}\cup S_1...\cup S_k$. Otherwise, suppose there exists a $B$ such that $A\subset B$ and $\{\phi\}\cup A_{i_1,1} \cup ...\cup A_{i_{k-1},k-1}\cup B$ is a consistent subset of $\{\phi\}\cup S_1...\cup S_k$, then $KB\cup B$ is consistent, which is a contradiction. Therefore, $A\in\{A_{ik}\}_{i=1,...,m}$. Next, we prove that for every $A_{ik}$, $KB\cup A_{ik}$ is a cardinality-maximal consistent subset of $KB\cup S_k$. It is clear that $KB\cup A_{ik}$ is consistent. Suppose there exists a subset $B$ of $S_k$ such that $A_{ik}\subset B$ and $KB\cup B$ is a consistent subset of $KB\cup S_k$, then by the proof in the "first" part, there must exist $A_{i_1,1}, ..., A_{i_{k-1},k-1}$ such that $\{\phi\}\cup A_{i_1,1} \cup ...\cup A_{i_{k-1},k-1}\cup B$ is a consistent subset of $\{\phi\}\cup S_1...\cup S_k$, where $A_{i_j,j}\in\{A_{ij}\}_{i=1,...,m}$. This is contradictory to the fact that $A_{ik} \in \{A_{ik}\}_{i=1,...,m}$. So we have proved that the set of cardinality-maximal consistent subsets of $KB\cup S_k$ is $\{KB\cup A_{ik}\}_{i=1,...,m}$. Therefore, the resulting knowledge base by applying CMR algorithm until the rank $k$ is $KB = \{\phi\} \cup \bigvee(\{A_{i1}\}_{i=1,...,m})\cup ..\cup \bigvee(\{A_{i,k}\}_{i=1,...,m})$. This finishes the induction.

By Lemma 1 and Lemma 2, we have the following theorem.

**Theorem 1** *Let $K = \{S_1, ..., S_n\}$ be a stratified knowledge base and $\phi$ a new formula. We then have the following equivalence:*

$$K_\phi \vdash_{Lex} \psi \quad iff \quad K_\phi \vdash_{CMR} \psi.$$

In (Benferhat et al. 2004), it has been shown that $K_\phi \vdash_{Lex} \psi$ iff $K_\phi \vdash_{DMA} \psi$. So we have the following corollary.

**Corollary 1** *Let $K$ be a stratified knowledge base and $\phi$ a formula. We then have the following equivalence:*

$$K_\phi \vdash_{DMA} \psi \quad iff \quad K_\phi \vdash_{CMR} \psi.$$

Now we will look at the complexity issues of CMR.

By Proposition 1, to resolve conflict information in each stratum, CMR needs in the worst case polynomial numbers of calls to SAT. So the DMA approach is computational harder than the CMR approach (under the usual assumption of complexity theory that $\Delta_2^p \subset \Sigma_p^2$ (Johnson 1990)).

However, the CMR approach can finish with a knowledge base with fewer formulae than the DMA approach.

**Lemma 3** *Let $S$ be a classical knowledge base and $\phi$ be a formula. Let $D$ be a conflict of $S\cup\{\phi\}$ and $S' = D\cap S$. Then $d_2(S')$ is consistent.*

**Lemma 4** *Let $S$ be a consistent classical knowledge base and $\phi$ be a formula. Let $\mathcal{C} = \{S'\subseteq S : \exists$ a conflict $C$ of $S\cup\{\phi\}$ such that $S' = C\cap S\}$. Suppose $S'\cap S'' = \emptyset$, for all $S', S''\subseteq S$, then $|S\circ_{CM}\phi| < |S\circ_{DMA}\phi|$.*

By Lemma 3 and Lemma 4 we have the following proposition.

**Proposition 4** *Let $K$ be a stratified knowledge base and $\phi$ be a new formula. Let $\psi_i$ be the formula obtained by the CMR algorithm in step $i$. Let $\mathcal{C}_i = \{S'_i\subseteq S_i : \exists$ a conflict $C_i$ of $S_i\cup\{\psi_{i-1}\}$ such that $S'_i = C_i\cap S_i\}$. Suppose for each $i$, if $\mathcal{C}_i \neq \emptyset$, then $S'_i\cap S''_i = \emptyset$, for all $S'_i, S''_i\subseteq S_i$, then $|\delta_{CMR}(K_\phi)| < |\delta_{DMA}(K_\phi)|$.*

**Example 5** *Let $K = \{S_1, S_2\}$ be such that $S_1 = \{c\vee d\vee e\}$ and $S_2 = \{\neg a, \neg b, \neg c, \neg d, \neg e\}$ and $\phi = a\vee b$. Let us apply the revision algorithm where $\circ = \circ_{CM}$. First we have $KB = \{a\vee b\}$. There is no conflict in $KB\cup S_1$, so $KB\leftarrow KB\cup S_1$. Let $\psi = (a\vee b)\wedge(c\vee d\vee e)$, then $S_2\bot\neg\psi = \{\{\neg a, \neg c, \neg d\}, \{\neg a, \neg c, \neg e\}, \{\neg a, \neg d, \neg e\}, \{\neg b, \neg c, \neg d\}, \{\neg b, \neg c, \neg e\}, \{\neg b, \neg d, \neg e\}\}$. So $\delta_{CMR}(K_\phi) = S_2\circ_{CM}\psi = \{\neg a\vee\neg b, a\vee b, c\vee d\vee e, \neg c \vee \neg d, \neg d \vee \neg e, \neg c \vee \neg e\}$. In contrast, suppose we apply the DMA algorithm. First we have $KB = \{\top\}$. There is no conflict in $KB\cup S_1$, so $KB\leftarrow KB\cup S_1$. Now, $S_2$ contradicts $KB$ due to the conflicts $\{a\vee b, \neg a, \neg b\}$ and $\{c\vee d\vee e, \neg c, \neg d, \neg e\}$. We now create all the possible pairwise disjunctions with $C = \{\neg a, \neg b, \neg c, \neg d, \neg e\}$: $d_2(C) = \{\neg a \vee \neg b, \neg a \vee \neg c, \neg a \vee \neg d, \neg a \vee \neg e, \neg b \vee \neg c, \neg b \vee \neg d, \neg b \vee \neg e, \neg c \vee \neg d, \neg c \vee \neg e, \neg d \vee \neg e\}$. Since $KB\cup d_2(C)$ is inconsistent, we create $d_3(C) = \{\neg a\vee\neg b\vee\neg c, \neg a\vee\neg b\vee\neg d, \neg a\vee\neg b\vee\neg e, \neg a\vee\neg c\vee\neg d, \neg a\vee\neg c\vee\neg e, \neg a\vee\neg d\vee\neg e, \neg b\vee\neg c\vee\neg d, \neg b\vee\neg c\vee\neg e, \neg b\vee\neg d\vee\neg e, \neg c\vee\neg d\vee\neg e\}$. Since $KB\cup d_3(C)$ is consistent, we add $d_3(C)$ to $KB$ and the algorithm stops. Then $\delta_{DMA}(K_\phi) = \{a\vee b, c\vee d\vee e, \neg a\vee\neg b\vee\neg c, \neg a\vee\neg b\vee\neg d, \neg a \vee \neg b \vee \neg e, \neg a \vee \neg c \vee \neg d, \neg a \vee \neg c \vee \neg e, \neg a \vee \neg d \vee \neg e, \neg b\vee\neg c\vee\neg d, \neg b\vee\neg c\vee\neg e, \neg b\vee\neg d\vee\neg e, \neg c\vee\neg d\vee\neg e\}$. $\delta_{DMA}(K_\phi)$ contains more formulas than $\delta_{CMR}(K_\phi)$.*

We also have the following lemma and proposition.

**Lemma 5** *Let $S$ be a consistent classical knowledge base and $\phi$ be a formula. Let $\mathcal{C} = \{S'\subseteq S : \exists$ a conflict $C$ of $S\cup\{\phi\}$ such that $S' = C\cap S\}$. Suppose $\cap(\mathcal{C}) \neq \emptyset$, then $|S\circ_{CM}\phi|\leq|S\circ_{DMA}\phi|$.*

**Proposition 5** *Let $K$ be a stratified knowledge base and $\phi$ be a new formula. Let $\psi_i$ be the formula obtained by the CMR algorithm in step $i$. Let $\mathcal{C}_i = \{S'_i\subseteq S_i : \exists$ a conflict $C_i$ of $S_i\cup\{\psi_{i-1}\}$ such that $S'_i = C_i\cap S_i\}$. Suppose for each $i$, if $\mathcal{C}_i \neq \emptyset$, then $\cap(\mathcal{C}_i) \neq \emptyset$, then $|\delta_{CMR}(K_\phi)|\leq|\delta_{DMA}(K_\phi)|$, but not vice verse.*

**Example 6** Let $K = \{a \vee b\} \cup \{\neg a, \neg b, \neg c \vee b, d\}$ and $\phi = c$. Let us apply the revision algorithm where $\circ = \circ_{CM}$. First we have $KB = \{c\}$. There is no conflict in $KB \cup S_1$, so $KB \leftarrow KB \cup S_1$. Let $\psi = c \wedge (a \vee b)$, then $S_2 \bot \neg \psi = \{\{\neg a, \neg c \vee b, d\}\}$. So $\delta_{CMR}(K_\phi) = S_2 \circ_{CM} \psi = \{c, \neg a, \neg c \vee b, d\}$. In contrast, suppose we apply the DMA algorithm. First we have $KB = \{c\}$. There is no conflict in $KB \cup S_1$, so $KB \leftarrow KB \cup S_1$. Now, $S_2$ contradicts $KB$ due to the conflicts $\{\{a \vee b, \neg a, \neg b\}, \{c, \neg b, \neg c \vee b\}\}$. We now create all the pairwise disjunctions with $C = \{\neg a, \neg b, \neg c \vee b\}$: $d_2(C) = \{\neg a \vee \neg b, \neg a \vee \neg c \vee b\}$. Since $KB \cup d_2(C)$ is consistent, we add $d_2(C)$ to $KB$ and the algorithm stops. Then $\delta_{DMA}(K_\phi) = \{c, \neg a \vee \neg b, \neg a \vee \neg c \vee b, d\}$. So $|\delta_{CMR}(K_\phi)| = |\delta_{DMA}(K_\phi)|$.

## 7 Conclusions

In this paper, we proposed a revision-based approach for conflict resolution. This approach has two different versions according to which class of revision operators are chosen. The first version is based on a model-based revision operator and the other is based on a formula-based revision operator. Our revision-based approach can be applied to deal with problems of belief revision and knowledge integration where the original knowledge base(s) is (are) stratified and inconsistent.

In this paper, we only considered two important revision operators, one is the Dalal's revision operator and the other is the cardinality-maximizing based revision operator. Since there are many other revision operators, an interesting problem is how to choose an appropriate revision operator. A suggestion is to compare the complexity of different revision methods and the sizes of the revised knowledge bases.

We have shown that the CMR approach can generate a knowledge base with fewer formulae than the DMA approach in some cases. An interesting question is, whether this conclusion still holds in general cases. This will be investigated in future work.

## 8 Acknowledge

We would like to thank the anonymous referees for their valuable comments to improve the quality of the paper.